\documentclass[conference]{IEEEtran}
\IEEEoverridecommandlockouts
\usepackage{cite}
\usepackage{amsmath,amssymb,amsfonts}
\usepackage{algorithmic}
\usepackage{graphicx}
\usepackage{textcomp}
\usepackage{xcolor}
\usepackage{todonotes}
\usepackage{newclude} 
\usepackage{orcidlink} 
\usepackage{pgfplots}

\def\BibTeX{{\rm B\kern-.05em{\sc i\kern-.025em b}\kern-.08em
    T\kern-.1667em\lower.7ex\hbox{E}\kern-.125emX}}
\begin{document}

\title{Interactive and Intelligent Root Cause Analysis in Manufacturing with Causal Bayesian Networks and Knowledge Graphs}

\author{\IEEEauthorblockN{1\textsuperscript{st} Christoph Wehner \orcidlink{0000-0003-0421-4113}}
\IEEEauthorblockA{\textit{Cognitive Systems Group} \\
\textit{University of Bamberg}\\
Bamberg, Germany \\
christoph.wehner@uni-bamberg.de 
} 
\and
\IEEEauthorblockN{2\textsuperscript{nd} Maximilian Kertel \orcidlink{0000-0003-3996-0642}
}
\IEEEauthorblockA{
\textit{Technology Development Battery Cell} \\
\textit{BMW Group}\\
Munich, Germany \\
maximilian.kertel@bmw.de} 
\and
\IEEEauthorblockN{3\textsuperscript{rd} Judith Wewerka \orcidlink{0000-0002-4809-2480}
}
\IEEEauthorblockA{
\textit{Digitalization and Innovation} \\
\textit{BMW Group}\\
Munich, Germany \\
judith.wewerka@bmw.de}
}

\maketitle

\begin{abstract}
\emph{Root Cause Analysis (RCA)} in the manufacturing of electric vehicles is the process of identifying fault causes. Traditionally, the \emph{RCA} is conducted manually, relying on process expert knowledge. Meanwhile, sensor networks collect significant amounts of data in the manufacturing process. Using this data for \emph{RCA} makes it more efficient. However, purely data-driven methods like \emph{Causal Bayesian Networks} have problems scaling to large-scale, real-world manufacturing processes due to the vast amount of potential cause-effect relationships (\emph{CER's}). Furthermore, purely data-driven methods have the potential to leave out already known \emph{CER's} or to learn spurious \emph{CER's}. The paper contributes by proposing an interactive and intelligent \emph{RCA} tool that combines expert knowledge of an electric vehicle manufacturing process and a data-driven machine learning method. It uses reasoning over a large-scale \emph{Knowledge Graph} of the manufacturing process while learning a \emph{Causal Bayesian Network}. In addition, an \emph{Interactive User Interface} enables a process expert to give feedback to the root cause graph by adding and removing information to the \emph{Knowledge Graph}.
The interactive and intelligent \emph{RCA} tool reduces the learning time of the \emph{Causal Bayesian Network} while decreasing the number of spurious \emph{CER's}. Thus, the interactive and intelligent \emph{RCA} tool closes the feedback loop between expert and machine learning method. 
\end{abstract}




\section{Introduction}
%
%
%
%

Machine learning is heavily used in driver assistant systems of electric vehicles \cite{wehner2022}. The electric vehicles benefit from machine learning not only on the road, but also during the manufacturing process as the latter has become more intelligent and data-driven.   
The manufacturing process is monitored via sensor networks, resulting in significant amounts of data hour by hour.
The data includes faults diagnosed by quality control process steps. However, detecting a fault in a complex manufacturing process does not necessarily uncover in which \emph{Process Steps} the fault was induced. Finding the cause-effect relationships \emph{(CER's)} leading to the diagnosed fault is called \emph{Root Cause Analysis (RCA)} \cite{oliveira2022}.

Traditionally, \emph{RCA} is a manual, labor-intense, and thus expensive process \cite{Serrat2017, Tay2008, Ruijters2015}, involving design of experiments to study manipulations and resulting effects. \emph{RCA} requires high amounts of process expert knowledge. Such that only process experts can conduct a \emph{RCA} in a reasonable timeframe. 

Digital support tools promise to make \emph{RCA} more efficient, data-driven, and less dependent on individual process knowledge \cite{oliveira2022}, to reduce costs and improve the performance of the manufacturing process. 

Previous work identified \emph{Causal Bayesian Networks} as a machine learning technique to automate \emph{RCA} \cite{pradhan2007} by learning \emph{CER's} between \emph{Sensor Variables}. The \emph{Causal Bayesian Networks} constructs a root cause graph over the manufacturing process, showing potential root causes \cite{michael2020}. 
However, learning \emph{Causal Bayesian Networks} for large manufacturing processes becomes prohibitively expensive as the search space for potential cause-effect relations explodes. 
This problem can be mitigated by including process expert knowledge in the learning process of the \emph{Causal Bayesian Networks} \cite{kertel2022}. 
In addition, expert knowledge can improve the learned root cause graph by identifying spurious \emph{CER's}.

The remaining challenges for scaling \emph{Causal Bayesian Networks} to large-scale manufacturing processes are to model process expert knowledge in detail and to provide an accessible way for the process expert to interact with and improve the root cause graph. The paper investigates those challenges by combining a large-scale \emph{Knowledge Graph} of the manufacturing process with a \emph{Causal Bayesian Network}. The \emph{Knowledge Graph} allows detailed modeling of large-scale manufacturing processes and enables the process expert to improve the root cause graph.  

In this work, a support tool for interactive and intelligent \emph{RCA} in the manufacturing process of electric vehicles is proposed. 
The \emph{RCA} is conducted by finding \emph{CER's} between \emph{Sensor Variables} of the manufacturing process. 

In detail, the paper contributes by:
\begin{itemize}
    \item Detailed modelling of process expert knowledge in a large-scale \emph{Knowledge Graph} for a real-world electric vehicle manufacturing process. 
    \item Automatically considering manufacturing knowledge from the \emph{Knowledge Graph} to drastically prune the search space while learning the \emph{Causal Bayesian Network}.
    \item Allowing the process expert to interact with and improve the root cause graph by explaining the \emph{Causal Bayesian Network} where and how to do better. 
\end{itemize}

\section{Related Work}
\emph{RCA} in manufacturing is dominated by methods defined in the ISO/IEC 31010 \cite{ISO}, including versions of the \emph{Five Why's} \cite{Serrat2017}, the \emph{Failure Mode and Effect Analysis (FMEA)} \cite{Tay2008}, or the \emph{Fault Tree Analysis} \cite{Ruijters2015}. 
The methods structure expert knowledge, such that a process expert can use it as a manual for \emph{RCA}. However, the decision-support tools do not take advantage of today's sensor networks and data.

Thus, machine learning methods were introduced in \emph{RCA}, enabling automated and intelligent decision-support tools \cite{MA2021107580, oliveira2022}. In particular, \emph{Causal Bayesian Networks} were identified as beneficial for \emph{RCA} by learning \emph{CER's} between measurements \cite{ALAEDDINI201111230, LOKRANTZ20181057, marazopoulou2016, LiKnowledgeDiscoveryBY}. 

For large and complex manufacturing processes, the number of possible \emph{Causal Bayesian Network} grows super-exponentially and its derivation becomes challenging \cite{ABELE20131843}. Additionally, one is not only interested in the existence of \emph{CER's}, but also in their strength to prioritize potential root causes. The derivation of the \emph{CER's} is called  \emph{structure learning}, while the identification of the strengths is called \emph{parameter learning}. \cite{michael2020} proposes an \emph{FMEA}-based approach, while \cite{pradhan2007} proposes an ontology of the manufacturing process to determine the structure of the \emph{Causal Bayesian Network}. However, the ontology only considers classes of entities and does not allow pruning the search space for individual measurements. Furthermore, in complex manufacturing processes many \emph{CER's} are unknown to the experts and both methods are unable to uncover them. On the contrary, \cite{ABELE20131843} combines expert knowledge and data to derive the \emph{Causal Bayesian Network}. This idea improves the quality of the root cause graph, as shown by \cite{WEIDL20051996}. 
While \emph{Sensor Variables} can be continuous, most of the aforementioned approaches consider a discretization to decrease the complexity of learning the \emph{Causal Bayesian Network} \cite{marazopoulou2016}. However, this practice may lead to such a degree of information loss that the data-driven identification of \emph{CER`s} becomes infeasible \cite{zhangFaultDiagnosis}. Recent approaches \cite{DAGsNoTearsNonparametric, CAM} for learning complex \emph{Causal Bayesian Networks} based on \emph{Structural Equation Models} (\emph{SEM's}) were applied on manufacturing processes \cite{kertel2022, DataFood} and the benefits of the inclusion of expert knowledge were demonstrated \cite{kertel2022}. 

In this work, we present a large-scale modeling of domain expertise into a \emph{Knowledge Graph} that leverages the sensor network topology for deriving the \emph{Causal Bayesian Network} of a real-world manufacturing process of electric vehicles. Furthermore, we show how the \emph{Knowledge Graphs} contributes to an interactive \emph{RCA} and closes the feedback loop for the \emph{Causal Bayesian Network} learning algorithm of \cite{kertel2022}.

\section{System Overview}
\begin{figure}[!ht]
  \centering
  \includegraphics[scale=0.25]{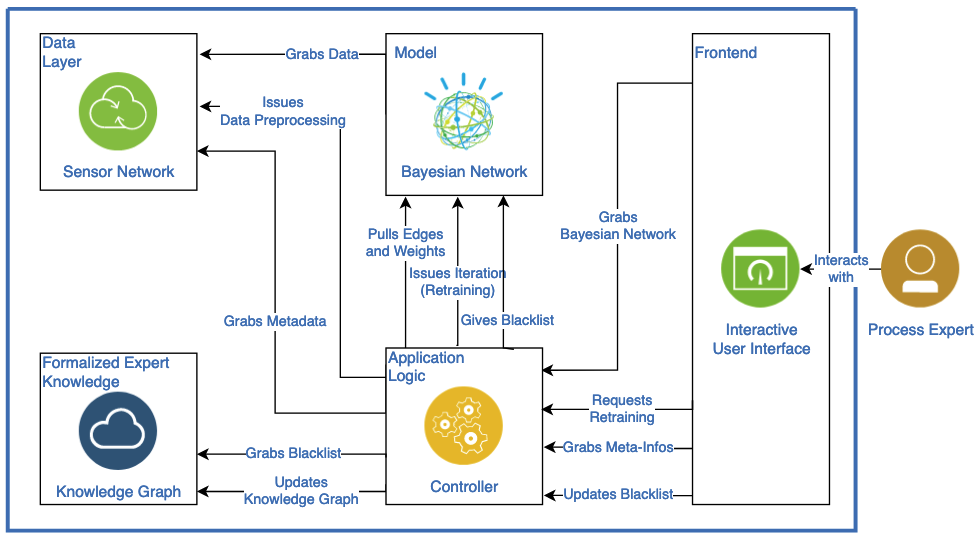}
  \caption{The system architecture of the interactive and intelligent root cause analysis tool.}\label{fig:system_architecture}
\end{figure}

The system for interactive and intelligent \emph{RCA} consists of five components (cf. Figure \ref{fig:system_architecture}), where each is a microservice on its own. 

The \emph{Controller} and the \emph{Data Layer} 
enable the core components.
The \emph{Knowledge Graph}, \emph{Causal Bayesian Network}, and \emph{Interactive User Interface} are the core components of the interactive and intelligent root cause analysis tool. 

The following describes the components' high-level functionalities and their interactions. Later on, the three core components are described in detail.

The heart of the system is a \emph{Controller} that handles the communication between all other components. It receives requests and triggers processes. 

The \emph{Data Layer} accesses data from a vehicle manufacturing-sensor network. Next, the data is preprocessed according to the input format of the \emph{Causal Bayesian Network} and stored in the \emph{Data Layer}. The preprocessing shall be elaborated in Subsection \ref{subsec:dataset}. 

Another primary source of data is the \emph{Knowledge Graph} (cf. Subsection \ref{subsec:knowledge_graph}).
The \emph{Knowledge Graph} holds formalized expert knowledge about the manufacturing process of electric vehicles, like the temporal-spatial relations of \emph{Stations}, \emph{Process Steps}, and \emph{Sensor Variables}.

The \emph{Causal Bayesian Network} takes the preprocessed sensor data from the \emph{Data Layer} for learning \emph{CER's} between the \emph{Sensor Variables} (cf. Subsection \ref{subsec:bayesian_network}). The information from the \emph{Knowledge Graph} is used while learning to reduce the search space drastically. The resulting \emph{Causal Bayesian Network} shows, which \emph{Sensor Variables} may induce faults in other \emph{Sensor Variables}. Thus, we call the graph learned by the \emph{Causal Bayesian Network} a root cause graph.

The root cause graph is visualized in an \emph{Interactive User Interface} (cf. Subsection \ref{subsec:ui}). The visualization enables experts to explore root causes in the manufacturing process of electric vehicles. Additionally, it allows for feedback of the process expert on the root cause graph. For example, the process expert may add an edge between two \emph{Sensor Variables} to the \emph{Knowledge Graph} or defines a \emph{Sensor Variable} as a \emph{Root Variable}. 





\subsection{Knowledge Graph}\label{subsec:knowledge_graph}
A \emph{Knowledge Graph} \cite{Hogan2021} is a directed labeled graph \cite{Harary1965, Gallian2001} and, therefore, optimal for storing highly relational data while preserving its semantics and allowing for deductive reasoning \cite{Hogan2021}. 

The \emph{Knowledge Graph} of the \emph{RCA} tool formalizes expert knowledge of the manufacturing process of electric vehicles. This allows the automatic, efficient, and large-scale mining of expert knowledge, which is then used while learning the \emph{Causal Bayesian Network}. 

\begin{figure}[h]
  \centering
  \includegraphics[scale=0.25]{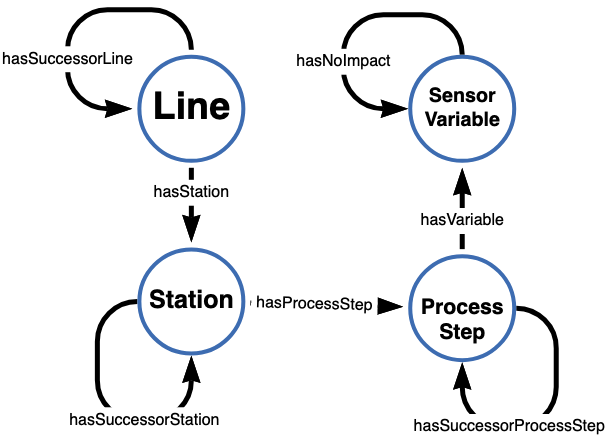}
  \caption{Schema of the manufacturing \emph{Knowledge Graph}. }\label{fig:root_finder_ontology}
\end{figure}

The expert knowledge of the manufacturing process is modeled as depicted in Figure \ref{fig:root_finder_ontology}. 

The \emph{Knowledge Graph} models the manufacturing process of electric vehicles as a sequence of \emph{Lines}. A \emph{Line} has various \emph{Stations}. The \emph{Stations} are also modeled as a sequence. Each \emph{Station} implements multiple \emph{Process Steps}. Again, the \emph{Knowledge Graph} models the sequence in which \emph{Process Steps} are executed. Each \emph{Process Step} measures \emph{Sensor Variables}. The \emph{Knowledge Graph} models explicitly a  \emph{``hasNoImpact"} relation between \emph{Sensor Variables} that are known to have no causal effect on each other. 

In addition, a \emph{Sensor Variable} can be a member of one subclass, which are \emph{Root}, \emph{Leaf}, and \emph{Irrelevant Variable}.

Formalizing expert knowledge of the manufacturing process in such a way allows automatic reasoning over the manufacturing process. The intelligent and interactive \emph{RCA} tool employs reasoning from the two following dimensions.


\begin{figure}[h]
  \centering
  \includegraphics[scale=0.21]{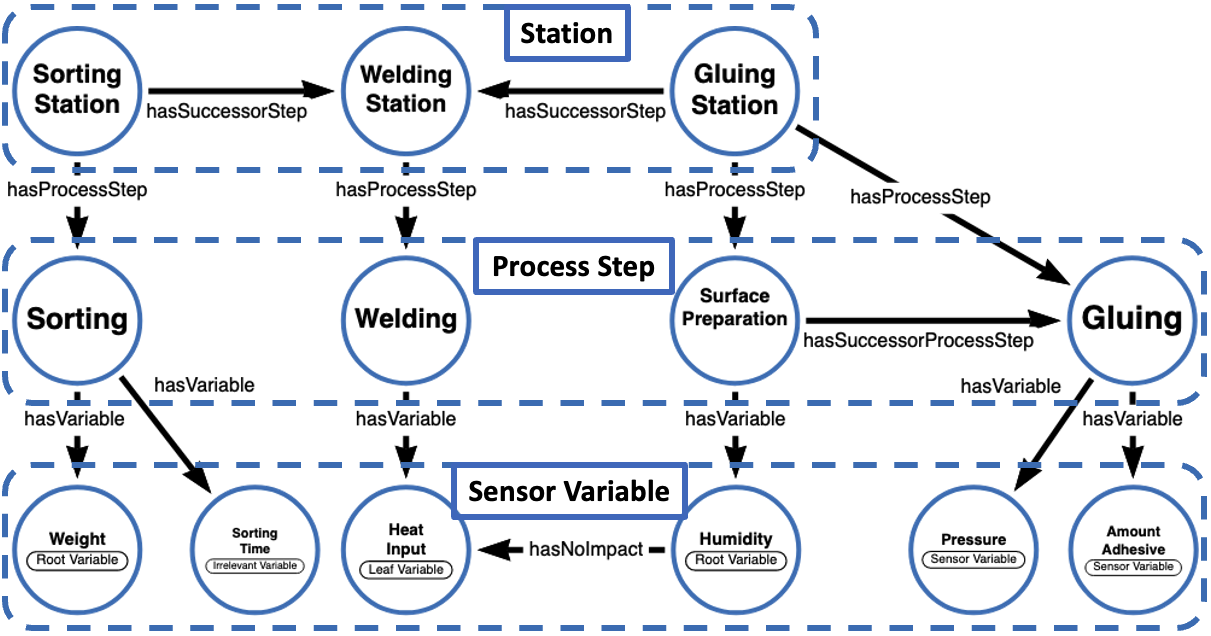}
  \caption{Example of the manufacturing \emph{Knowledge Graph}. }\label{fig:example_kg}
\end{figure}

\paragraph{Temporal-spatial relations between Variables} 
The \emph{Line} is modeled as a sequence of \emph{Stations}, each with a sequence of \emph{Process Steps}. At the bottom of this hierarchy are sensors measuring the \emph{Sensor Variables}. Thus, the \emph{Knowledge Graph} constructs a topology of the sensor network. It enables automatic reasoning to create a partial ordering over the \emph{Sensor Variables}. For every \emph{Sensor Variable}, it is known which \emph{Sensor Variable} is measured before and after and a \emph{Sensor Variable} measured in an earlier \emph{Process Step} may influence a \emph{Sensor Variable} measured in a subsequent \emph{Process Step}, but not vice versa.

Let us consider the \emph{Knowledge Graph} from Figure \ref{fig:example_kg}. The topology of the sensor network allows inferring that the \emph{Sensor Variables} \emph{``Weight"} and \emph{``Sorting Time"} may causally impact the \emph{Sensor Variable} \emph{``Heat Input"}, but not vice versa and not the \emph{Sensor Variables} \emph{``Humidity"}, \emph{``Amount Adhesive"}, and \emph{``Pressure"}.

\begin{figure*}[!tbp]
  \centering
  \begin{minipage}[b]{0.25\textwidth}
    \includegraphics[width=\textwidth]{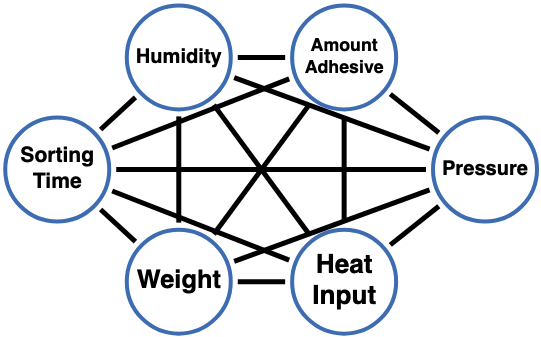}
    \caption{The potential \emph{CER's} between the \emph{Sensor Variables} of the manufacturing \emph{Knowledge Graph} (cf. Figure \ref{fig:example_kg}) without considering the topology of the sensor network.}
    \label{fig:causal_relations_without_knowledge}
  \end{minipage}
  \hfill
  \begin{minipage}[b]{0.25\textwidth}
    \includegraphics[width=\textwidth]{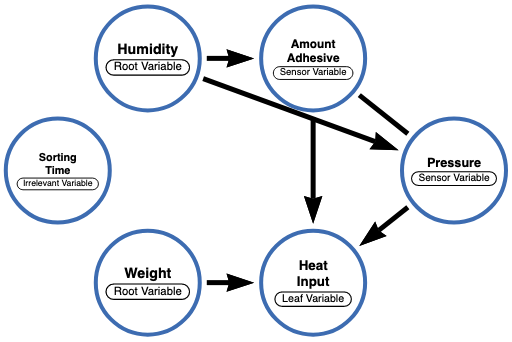}
    \caption{The potential \emph{CER's} between the \emph{Sensor Variables} of the manufacturing \emph{Knowledge Graph} (cf. Figure \ref{fig:example_kg}) while considering the topology of the sensor network.}
    \label{fig:causal_relations_with_knowledge}
  \end{minipage}
\hfill
\begin{minipage}[b]{0.25\textwidth}
\includegraphics[width=\textwidth]{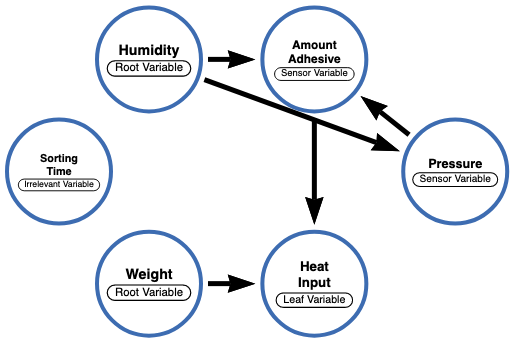}
  \caption{The assumed true causal graph between the \emph{Sensor Variables}. Its edges are contained in the relations of the potential \emph{CER's} graph of Figure \ref{fig:causal_relations_with_knowledge}.}\label{fig:true_causal_graph}
\end{minipage}
\end{figure*}

The example shows the reasoning on how a significant amount of possible \emph{CER's} between \emph{Sensor Variables} are excluded a priori and do not have to be considered while training the \emph{Causal Bayesian Network}. This reduces the search space of the \emph{Causal Bayesian Network} learning algorithm drastically, compared to a naive approach of learning the \emph{Causal Bayesian Network} solely on tabular data (cf. Figures \ref{fig:causal_relations_without_knowledge} and \ref{fig:causal_relations_with_knowledge}). 

\paragraph{Properties of Variable subclasses}

Additionally, the \emph{Knowledge Graph} incorporates expert knowledge over the \emph{Sensor Variables} in the form of their subclass membership. This information is introduced to the \emph{Knowledge Graph} by the feedback of a process expert and is vital to reduce the \emph{Causal Bayesian Network}'s learning algorithm's search space.

\emph{Root Variables} cannot be impacted by any \emph{Sensor Variables}. However, they may impact other \emph{Sensor Variables}. For example, the \emph{Sensor Variable} \emph{``Humidity"} in Figure \ref{fig:example_kg} cannot be affected by \emph{Sensor Variables}, as the air humidity is external to the manufacturing process. However, \emph{``Humidity"} may affect other \emph{Sensor Variables} in the manufacturing process. 

\emph{Leaf Variables} are the inverse of \emph{Root Variables}. They may be impacted by \emph{Sensor Variables}, but cannot impact other \emph{Sensor Variables}. The \emph{``Heat Input"} \emph{Sensor Variable} in Figure \ref{fig:example_kg} is an example of a \emph{Leaf Variable}, as it is the only \emph{Sensor Variable} of the final \emph{Process Step}. 

Finally, there are \emph{Irrelevant Variables}. Any manufacturing process measures a minor amount of \emph{Sensor Variables} that do not impact any other \emph{Sensor Variables} in the process. \emph{Irrelevant Variables} are artifacts of unclear requirements, highly specific use cases, or legislation.
The \emph{``Sorting Time"} \emph{Sensor Variable} in Figure \ref{fig:example_kg} is an example, as the sorting time does not impact other \emph{Sensor Variables}. Thus, they shall be excluded from the \emph{RCA}.  

Figures \ref{fig:causal_relations_without_knowledge} and \ref{fig:causal_relations_with_knowledge} show the impact of reasoning over the temporal-spatial relations between \emph{Sensor Variables} and their subclasses on the search space of the \emph{Causal Bayesian Network} learning algorithm. 
In Figure \ref{fig:causal_relations_without_knowledge}, every \emph{Sensor Variable}`s fault is potentially caused by any other \emph{Sensor Variable}, as the sensor network's topology modeled in the \emph{Knowledge Graph} is not considered. This amounts to a total of 30 possible \emph{CER's} between the six \emph{Sensor Variables}. 
However, considering the sensor network topology, the search space of the \emph{Causal Bayesian Network} learning algorithm is reduced to seven potential \emph{CER's} between the \emph{Sensor Variables} (cf. Figure \ref{fig:causal_relations_with_knowledge}). 
This example shows that automated reasoning over the expert knowledge formalized in the \emph{Knowledge Graph} prunes the search space of the \emph{Causal Bayesian Network} learning algorithm significantly.

\subsection{Causal Bayesian Networks} 
\label{subsec:bayesian_network}
This section introduces important concepts of \emph{Causal Bayesian Networks}, which represent \emph{CER's} behind data sets, and how it incorporates the \emph{Knowledge Graph} into its training. 
\emph{CER's} are dependencies, where the manipulation of one \emph{Sensor Variable} changes the value of the other even if all remaining \emph{Sensor Variables} are fixed. 
Assume the true causal graph is depicted in Figure \ref{fig:true_causal_graph}. Then a change of \emph{``Pressure"} influences \emph{``Heat Input"} through the path \emph{``Pressure"} $\rightarrow$ \emph{``Amount Adhesive"} $\rightarrow$ \emph{``Heat Input"}. For example, a higher \emph{``Pressure"} might lead to a higher \emph{``Amount Adhesive"}, which in turn results in a larger \emph{``Heat Input"}. Now imagine that after an increase of \emph{``Amount Adhesive"} we manually change all \emph{Sensor Variables} but \emph{``Heat Input"} back to their original level. Then, the manipulation of \emph{``Pressure"} does not impact \emph{``Heat Input"} anymore, as it is only directly influenced by \emph{``Weight"} and \emph{``Amount Adhesive"}, which are as before. Therefore, the influential path \emph{``Pressure"} $\rightarrow$ \emph{``Amount Adhesive"} $\rightarrow$ \emph{``Heat Input"} is blocked. 
We understand these direct impacts as the \emph{CER's}. They are pivotal for root-cause analysis, process control, and process understanding. 
\emph{CER's} imply an order over the \emph{Sensor Variables}, which we call the \emph{causal order}. In our running example (cf. Figure \ref{fig:true_causal_graph}), one possible causal order of the \emph{Sensor Variables} is (\emph{``Sorting Time"}, \emph{``Weight"} \emph{``Humidity"}, \emph{``Pressure"}, \emph{``Amount Adhesive"}, \emph{``Heat Input"}).

This paper's scope is to derive the unknown causal graph. 
Instead of labor and cost-intensive design of experiments, we would like to leverage our sensor network along the manufacturing process to derive the causal graph in a data-driven manner. However, if we rely exclusively on data, this is challenging due to the following reasons: 
\begin{enumerate}
    \item \textbf{Confounding:} Even if we have identified a correlation, say between \emph{``Amount Adhesive"} and \emph{``Heat Input"}, it might be, that there is a third \emph{Sensor Variable}, say \emph{``Machine Operator"}, impacting both and there is no direct impact between the two. We call this third \emph{Sensor Variable} a confounder. Besides pathological cases, confounders lead to extra spurious relationships. 
    \item \textbf{Causal Order:} Even if we can exclude confounding it remains still unclear, whether \emph{``Amount Adhesive"} is the cause and \emph{``Heat Input"} the effect or it is vice versa.
    \item \textbf{Complexity:} The number of potential graphs grows super-exponentially. Beyond a small number of \emph{Sensor Variables}, it is thus impossible to evaluate all of them. 
    This is especially worrisome if there are many cause-effect pairs.
\end{enumerate}

\subsubsection*{Expert Knowledge to the Rescue}
In the following, we present how expert knowledge contributes to derive the causal graph. As Section \ref{subsec:knowledge_graph} mentions, the number of potential causal graphs can be drastically reduced by the sensor network's topology and expert knowledge on the manufacturing process:
\begin{enumerate}
    \item sensor network topology provides a \textbf{partial ordering} of the \emph{Sensor Variables} and
    \item expert knowledge helps to exclude certain relationships and thus \textbf{avoids spurious relationships}. 
\end{enumerate}
Additionally, 
even a partial ordering drastically reduces the number of potential graphs. The classification of \emph{Sensor Variables} into roots and leafs reduces not only the number of potential edges but also the number of relevant orderings. For example, as \emph{``Weight"} and \emph{``Humidity"} are defined as \emph{Root Variables}, the relevant orderings are those that assign them to any of the first two positions. This reduces the number of permutations by a factor of $30$. The effect of \emph{Leaf Variables} is analogous. In the following, we present how we leverage the background information for causal graph identification.
\subsubsection*{Causal Additive Models}
The derivation of the \emph{CER's} from observed data is of central interest in many domains\cite{wille2004sparse, saxe2020computational}. 
For this task, score-based methods relying on \emph{Structural Equation Models (SEM's)} \cite{PetersBook} have recently become increasingly popular due to their ability to incorporate machine learning methods\cite{DAGsNoTearsNonparametric, CAM, DirectLiNGAM, DAGGNN}. The underlying assumption is that all \emph{Sensor Variables} can be expressed as a function of inputs and a noise term, that captures unknown influences. In the example of Figure \ref{fig:causal_relations_with_knowledge}, \emph{``Heat Input"} can be described by Equation  \ref{eq:sem}.
\begin{equation}
\label{eq:sem}
\mathit{Heat Input} = f(\mathit{Weight}, \mathit{Amount Adhesive},\mathit{Noise}) 
\end{equation}
We emphasize that this follows an intuitive understanding of a manufacturing process. \emph{Process Steps} transform input material to an output, while they are impacted by machine settings and the environment. 
We assume that the data follows a \emph{Causal Additive Model}, where Equation (\ref{eq:sem}) is replaced by: 
\begin{equation}
\mathit{Heat Input} = f_1(\mathit{Humidity}) + f_2(\mathit{Amount Adhesive}) + \mathit{Noise}
\end{equation}
The \emph{Noise} is normally distributed and $f_1$ and $f_2$ are non-linear. 
Under mild assumptions \cite{PetersBook} on $f_1$ and $f_2$, one can derive the true causal graph from observed data. \cite{CAM} proposes an approach, that identifies the graph using the following steps: 
\begin{enumerate}
    \item Find the causal ordering of the \emph{Sensor Variables}
    \item Identify the \emph{CER's}.
\end{enumerate}
Given $N$ observations of $p$ \emph{Sensor Variables} by $\left(x_{\ell 1}, \ldots, x_{\ell p}\right), 1\leq \ell \leq N$. To derive the causal ordering in step (1), search for the permutation $\pi$ on $\{1, \ldots, p\}$ such that Equation \ref{eq:optimization_goal} is minimized and $\widehat{f^\pi_{k j}}$ are learned by maximum likelihood estimation \cite{CAM}. 
\begin{equation} 
\label{eq:optimization_goal}
S(\pi) := \sum_{k = 1 }^p  \log \left[ \sum_{\ell=1}^N \left(  x_{\ell k} - \sum_{\pi(j) < \pi(k)} \widehat{f^\pi_{k j}}(x_{\ell  j})\right)^2 \right],
\end{equation}
If the \emph{Knowledge Graph} provides a complete ordering, then we skip the first step, and the graph identification is found using regression techniques \cite{ShojaieKnownOrdering}. Unfortunately, the ordering is usually partial, as \emph{Sensor Variables} measured at the same station cannot be ordered. We employ the algorithm of \cite{kertel2022}, which proposes an efficient adaption in case of expert knowledge. It limits step (1) to finding the causal ordering of the \emph{Sensor Variables} within the \emph{Process Steps} while mining the rest of the ordering from the \emph{Knowledge Graph}

\subsection{Interactive User Interface} \label{subsec:ui}

\begin{figure*}[h]
  \centering
  \includegraphics[scale=0.27]{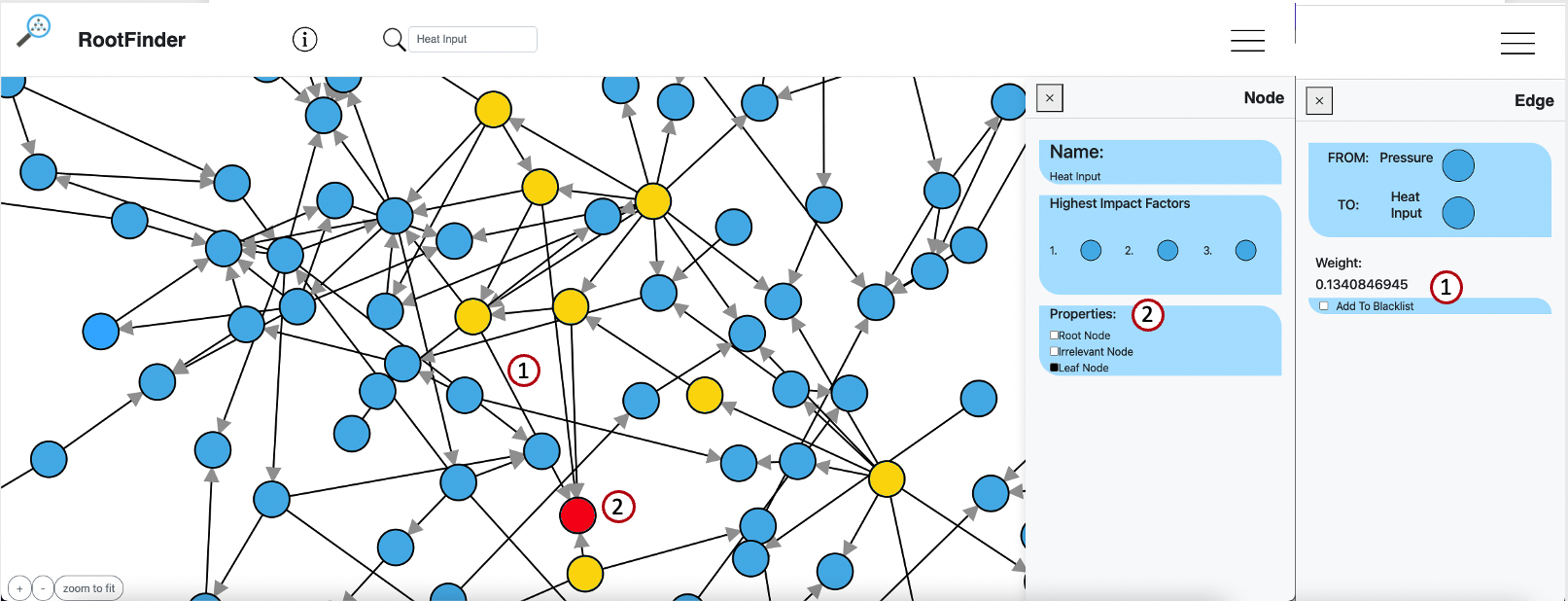}
  \caption{The user interface of the interactive and intelligent root cause analysis tool. The fault is marked red. \emph{Sensor Variables} part of the learned root cause path are marked yellow. The numbers in the red circle highlight the interactions.}\label{fig:screenshot}
\end{figure*}

The \emph{Interactive User Interface} visualizes the learned \emph{Causal Bayesian Network} (cf. Figure \ref{fig:screenshot}). 
Thereby, the \emph{Interactive User Interface} enables no-code usage and adaption of the root cause graph by a process expert. 

\paragraph{Root Cause Analysis}
The workflow supported by the \emph{Interactive User Interface} is as follows (cf. Figure \ref{fig:screenshot}): 

The process expert searches for the \emph{Sensor Variable} with the diagnosed fault.
All possible root cause paths for the faulty \emph{Sensor Variable} are displayed to the process expert. 
The process expert now sees the information on which other \emph{Sensor Variables} the faulty \emph{Sensor Variable} depends.
That way, without further labor-intense analysis, the process expert can move to the physical manufacturing process and check all \emph{Process Steps} related to the identified \emph{Sensor Variables}. The process expert benefits from a highly directed root cause search in the physical manufacturing process. Thus, the interactive and intelligent root cause analysis tool minimizes and sometimes even prevents the downtime of the manufacturing process and maximizes the manufacturing process's output.

\paragraph{Expert Feedback} 
The \emph{RCA} is supported by various means of interaction with the root cause graph (cf. Figure \ref{fig:screenshot}).

The process expert can choose the \emph{Product} to be considered in the \emph{RCA}. This results in accurate root cause graphs, as \emph{Sensor Variable} measurements significantly depend on the \emph{Product} measured. In addition, the process expert can select the \emph{RCA} timeframe via the \emph{Interactive User Interface}. Selecting a timeframe enables the process expert to look precisely at the period when the faults were detected. It thus customizes the \emph{RCA} to the individual fault at hand. 

However, the root cause graph might include spurious \emph{CER's} and needs to be corrected. In this case, the expert is able to explain the \emph{Causal Bayesian Network} where and how to do better, via adding or removing information to the \emph{Knowledge Graph}. 
The \emph{Causal Bayesian Network} may learn a \emph{CER} between two \emph{Sensor Variables}, which has to be rejected based on the knowledge of the process expert. The process expert can select such an edge and check "Add to Blacklist", which adds an \emph{``hasNoImpact"} relation between the two \emph{Sensor Variables} to the \emph{Knowledge Graph} (cf. Figure \ref{fig:screenshot} (1)). 
The expert feedback is considered at the next iterations of learning the \emph{Causal Bayesian Network}, and thus is accounted for in the next \emph{RCA's}. 

In addition, spurious \emph{CER's} might be the cause of incomplete information over subclass membership of \emph{Sensor Variables} in the \emph{Knowledge Graph}. Thus, the process expert can add and remove \emph{Root}, \emph{Leaf}, and \emph{Irrelevant Variable} subclass membership from \emph{Sensor Variables} (cf. Figure \ref{fig:screenshot} (2)). This feedback is also considered in the next iteration of the \emph{Causal Bayesian Network}.
Assuming that the expert feedback is correct, the root cause graph converges each iteration closer to the true root cause graph.  
\section{Evaluation} 
The evaluation is conducted on a real-world electric vehicle manufacturing process. One \emph{Line} is taken as an example to show the \emph{RCA} with the interactive and intelligent tool.

\subsection{Data}  \label{subsec:dataset}
The data is two-folded. There is the \emph{Knowledge Graph} and the \emph{Data Layer} as a source of data. 

The \emph{Knowledge Graph} is implemented in \emph{Neo4J} \cite{lal2015}. This allows querying all knowledge via the Cypher API. Cypher is a deductive reasoning query language \cite{lal2015}. 
The real-world \emph{Knowledge Graph} holds a total of 100,015 nodes and 417,944 relationships. This amount of expert knowledge constitutes a large-scale knowledge graph. For the \emph{Line} of this evaluation, there are 53 \emph{Stations} with 96 \emph{Process Steps} and 1683 \emph{Sensor Variables} and a total of 2143 relations modeled. 

To learn the \emph{Causal Bayesian Network} we prepare the data by assigning the \emph{Sensor Variables} to individual output products. Then, we preprocess the data by removing columns with only one value. Further, we remove one part of a column pair, if they have a correlation above $0.95$ to avoid collinearity. Afterwards, we remove columns and rows, where more than $50\%$ of the values are missing. As the rate of missing values is low, we simply impute the column's mean for missing values. 
The presented approach is generic and can be applied dynamically to different products and time windows. For this section, we consider data of one day and for one specific product type. 
\subsection{Experiments} \label{subsec:experiments}
The contribution of the paper is to give a real-world example of how to interactively improve the performance of \emph{Causal Bayesian Networks} for \emph{RCA} with large-scale modelled expert knowledge in the form of a \emph{Knowledge Graph}. 
The contribution is shown by three experiments.

\begin{figure}
\centering
\begin{tikzpicture}[scale=0.66]
\begin{axis}[
    xlabel={Expert Knowledge in Percent (\%)},
    ylabel = Number (n),
    ymin=300, ymax=1100,
    xmin=15, xmax=110,
    ylabel=Number (n),
    xtick={0,25,50,75,100},
    axis y line*=left,
    legend pos=north west,
    ymajorgrids=true,
    grid style=dashed,
]


\addplot[
    color=red,
    mark=square,
    ]
coordinates {
    (100,398)(75,955)(50,1012)(25,974)
    };        
\end{axis}
\end{tikzpicture}
\caption{Decrease of learning time and \emph{CER's} with an increase of data from the \emph{Knowledge Graph}.}
\label{dia:evaluation}
\end{figure}
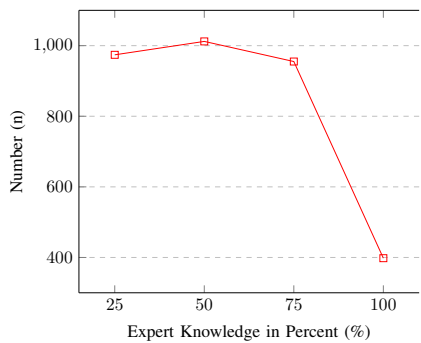

\subsubsection{Decrease in learning time of the Causal Bayesian Network by including the Knowledge Graph} \label{subsubsec:experment_time}
As argued above, one major motivation for including expert knowledge is a decrease in the learning time of the \emph{Causal Bayesian Network} algorithm.
Figure \ref{dia:evaluation} shows the average decrease in learning time with an increase in expert knowledge. The experiment was conducted five times. For each run-through, randomly 25\%, 50\%, 75\%, and 100\% of data from the \emph{Knowledge Graph} were selected for training the \emph{Causal Bayesian Network}. A decline by $79.0\%$ in training time can be observed, with an increase in data from the Knowledge Graph. Thus, expert knowledge optimizes and enables the training of \emph{Causal Bayesian Network} in large-scale manufacturing processes.

\subsubsection{Decline of spurious CER's in the Causal Bayesian Network by including the Knowledge Graph}
As argued above, \emph{Causal Bayesian Networks} for manufacturing processes tend to include spurious \emph{CER's}.
Figure \ref{dia:evaluation} shows the average decline in spurious CER's with an increase in expert knowledge. The experiment was conducted as described in \ref{subsubsec:experment_time}. A steep decline in the amount of learned \emph{CER's} is observable. The result indicates that data from the \emph{Knowledge Graph} prunes $60.6\%$ of spurious \emph{CER's}. This leads to a more accurate root cause graph and thus increases the performance of the root cause analysis tool.
The following experiment shall underline the latter claim.  

\subsubsection{Expert feedback improves the Causal Bayesian Network} 
To evaluate the usefulness of the interactive component, we extend the evaluation proposed by \cite{kertel2022}.
A learned \emph{Causal Bayesian Network} is compared to a partially known root cause graph, using an \emph{adapted Structural Hamming Distance (aSHD)}, which describes the deviation of the learned \emph{Causal Bayesian Network} from the partially known root cause graph \cite{kertel2022}. A lower number indicates a better result. The learning algorithm is applied on $100$ random samples of eight \emph{Sensor Variables}, using the \emph{temporal-spatial ordering}. For each learned \emph{Causal Bayesian Network}, the \emph{aSHD} is calculated. The mean and standard deviation of the \emph{aSHD} are $0.47$ and $0.74$, respectively. If an expert now adds one \emph{hasNoImpact} relation to the \emph{Knowledge Graph}, the mean and standard deviation of the \emph{aSHD} go down to $0.44$ and $0.70$. This corresponds to an improvement of the \emph{aSHD} by almost $6\%$ and the learning algorithm is stabilized, as the standard deviation is lower. The results illustrate the benefits of including expert knowledge and exemplify one iteration step of the interactive and intelligent root cause analysis tool.
\section{Conclusion}\label{sec:conclusion}
This work proposes an interactive and intelligent root cause analysis tool for the manufacturing process of electric vehicles. It shows how to model detailed expert knowledge of a real-world manufacturing process in a large-scale \emph{Knowledge Graph}, which is used for automatic reasoning over potential root causes. It is described how the reasoning is used to prune the search space of a \emph{Causal Bayesian Network} learning algorithm. In addition, this work shows how to include expert feedback in the learning of the \emph{Causal Bayesian Network} via the \emph{Interactive User Interface} and the \emph{Knowledge Graph} to identify and exclude spurious \emph{CER's}. The evaluation of the interactive and intelligent root cause analysis tool on a real-world manufacturing process of electric vehicles proves the tool's feasibility. 

Machine learning makes \emph{RCA} more efficient. However, it will only succeed if we rely on data and expert knowledge. Let us put the human back in the loop. 

\section*{Acknowledgment}
This research was co-funded by the Bavarian Ministry of Economic Affairs, Regional Development and Energy, funding line "Digitization" of the BayVFP, project KIProQua, and the BMW Group. In particular, we thank Joscha Eirich for enabling this research and for his feedback on the intelligent and interactive root cause analysis tool. 

\bibliographystyle{IEEEtran}
\bibliography{conference_101719}

\begin{thebibliography}{10}
\providecommand{\url}[1]{#1}
\csname url@samestyle\endcsname
\providecommand{\newblock}{\relax}
\providecommand{\bibinfo}[2]{#2}
\providecommand{\BIBentrySTDinterwordspacing}{\spaceskip=0pt\relax}
\providecommand{\BIBentryALTinterwordstretchfactor}{4}
\providecommand{\BIBentryALTinterwordspacing}{\spaceskip=\fontdimen2\font plus
\BIBentryALTinterwordstretchfactor\fontdimen3\font minus
  \fontdimen4\font\relax}
\providecommand{\BIBforeignlanguage}[2]{{%
\expandafter\ifx\csname l@#1\endcsname\relax
\typeout{** WARNING: IEEEtran.bst: No hyphenation pattern has been}%
\typeout{** loaded for the language `#1'. Using the pattern for}%
\typeout{** the default language instead.}%
\else
\language=\csname l@#1\endcsname
\fi
#2}}
\providecommand{\BIBdecl}{\relax}
\BIBdecl

\bibitem{wehner2022}
C.~Wehner, F.~Powlesland, B.~Altakrouri, and U.~Schmid, ``Explainable online
  lane change predictions on a digital twin with a layer normalized lstm
  and layer-wise relevance propagation,'' in \emph{Advances and Trends in
  Artificial Intelligence. Theory and Practices in Artificial Intelligence},
  H.~Fujita, P.~Fournier-Viger, M.~Ali, and Y.~Wang, Eds.\hskip 1em plus 0.5em
  minus 0.4em\relax Cham: Springer International Publishing, 2022, pp.
  621--632.

\bibitem{oliveira2022}
\BIBentryALTinterwordspacing
E.~e~Oliveira, V.~L. Miguéis, and J.~L. Borges, ``Automatic root cause
  analysis in manufacturing: an overview and conceptualization,'' \emph{Journal
  of Intelligent Manufacturing}, 2022. [Online]. Available:
  \url{https://doi.org/10.1007/s10845-022-01914-3}
\BIBentrySTDinterwordspacing

\bibitem{Serrat2017}
O.~Serrat, \emph{The Five Whys Technique}.\hskip 1em plus 0.5em minus
  0.4em\relax Singapore: Springer Singapore, 2017, pp. 307--310.

\bibitem{Tay2008}
\BIBentryALTinterwordspacing
K.~M. Tay and C.~P. Lim, ``On the use of fuzzy inference techniques in
  assessment models: part ii: industrial applications,'' \emph{Fuzzy
  Optimization and Decision Making}, vol.~7, pp. 283--302, 2008. [Online].
  Available: \url{https://doi.org/10.1007/s10700-008-9037-y}
\BIBentrySTDinterwordspacing

\bibitem{Ruijters2015}
\BIBentryALTinterwordspacing
E.~Ruijters and M.~Stoelinga, ``Fault tree analysis: A survey of the
  state-of-the-art in modeling, analysis and tools,'' \emph{Computer Science
  Review}, vol. 15-16, pp. 29--62, 2015. [Online]. Available:
  \url{https://www.sciencedirect.com/science/article/pii/S1574013715000027}
\BIBentrySTDinterwordspacing

\bibitem{pradhan2007}
S.~Pradhan, R.~Singh, K.~Kachru, and S.~Narasimhamurthy, ``A bayesian network
  based approach for root-cause-analysis in manufacturing process,'' in
  \emph{2007 International Conference on Computational Intelligence and
  Security (CIS 2007)}, 2007, pp. 10--14.

\bibitem{michael2020}
M.~Kirchhof, K.~Haas, T.~Kornas, S.~Thiede, M.~Hirz, and C.~Hermann, ``Root
  cause analysis in lithium-ion battery production with fmea-based large-scale
  bayesian network,'' 06 2020.

\bibitem{kertel2022}
\BIBentryALTinterwordspacing
M.~Kertel, S.~Harmeling, and M.~Pauly, ``Learning causal graphs in
  manufacturing domains using structural equation models,'' 2022. [Online].
  Available: \url{https://arxiv.org/abs/2210.14573}
\BIBentrySTDinterwordspacing

\bibitem{ISO}
ISO/IEC, ``{Risk management — Risk assessment techniques},'' International
  Organization for Standardization, Geneva, CH, Standard, Jun. 2019.

\bibitem{MA2021107580}
\BIBentryALTinterwordspacing
Q.~Ma, H.~Li, and A.~Thorstenson, ``A big data-driven root cause analysis
  system: Application of machine learning in quality problem solving,''
  \emph{Computers and Industrial Engineering}, vol. 160, p. 107580, 2021.
  [Online]. Available:
  \url{https://www.sciencedirect.com/science/article/pii/S0360835221004848}
\BIBentrySTDinterwordspacing

\bibitem{ALAEDDINI201111230}
\BIBentryALTinterwordspacing
A.~Alaeddini and I.~Dogan, ``Using bayesian networks for root cause analysis in
  statistical process control,'' \emph{Expert Systems with Applications},
  vol.~38, no.~9, pp. 11\,230--11\,243, 2011. [Online]. Available:
  \url{https://www.sciencedirect.com/science/article/pii/S0957417411003952}
\BIBentrySTDinterwordspacing

\bibitem{LOKRANTZ20181057}
\BIBentryALTinterwordspacing
A.~Lokrantz, E.~Gustavsson, and M.~Jirstrand, ``Root cause analysis of failures
  and quality deviations in manufacturing using machine learning,''
  \emph{Procedia CIRP}, vol.~72, pp. 1057--1062, 2018, 51st CIRP Conference on
  Manufacturing Systems. [Online]. Available:
  \url{https://www.sciencedirect.com/science/article/pii/S2212827118303895}
\BIBentrySTDinterwordspacing

\bibitem{marazopoulou2016}
K.~Marazopoulou, R.~Ghosh, P.~Lade, and D.~Jensen, ``Causal discovery for
  manufacturing domains,'' 05 2016.

\bibitem{LiKnowledgeDiscoveryBY}
\BIBentryALTinterwordspacing
J.~Li and J.~Shi, ``Knowledge discovery from observational data for process
  control using causal bayesian networks,'' \emph{IIE Transactions}, vol.~39,
  no.~6, pp. 681--690, 2007. [Online]. Available:
  \url{https://doi.org/10.1080/07408170600899532}
\BIBentrySTDinterwordspacing

\bibitem{ABELE20131843}
\BIBentryALTinterwordspacing
L.~Abele, M.~Anic, T.~Gutmann, J.~Folmer, M.~Kleinsteuber, and B.~Vogel-Heuser,
  ``Combining knowledge modeling and machine learning for alarm root cause
  analysis,'' \emph{IFAC Proceedings Volumes}, vol.~46, no.~9, pp. 1843--1848,
  2013, 7th IFAC Conference on Manufacturing Modelling, Management, and
  Control. [Online]. Available:
  \url{https://www.sciencedirect.com/science/article/pii/S1474667016345633}
\BIBentrySTDinterwordspacing

\bibitem{WEIDL20051996}
\BIBentryALTinterwordspacing
G.~Weidl, A.~Madsen, and S.~Israelson, ``Applications of object-oriented
  bayesian networks for condition monitoring, root cause analysis and decision
  support on operation of complex continuous processes,'' \emph{Computers and
  Chemical Engineering}, vol.~29, no.~9, pp. 1996--2009, 2005. [Online].
  Available:
  \url{https://www.sciencedirect.com/science/article/pii/S009813540500133X}
\BIBentrySTDinterwordspacing

\bibitem{zhangFaultDiagnosis}
D.~Zhang, L.~Wang, Q.~Hong, and K.~Zhang, ``Research on fault diagnosis of
  steam turbine based on bayesian network,'' \emph{Journal of Physics:
  Conference Series}, vol. 1754, p. 012136, 02 2021.

\bibitem{DAGsNoTearsNonparametric}
X.~Zheng, C.~Dan, B.~Aragam, P.~Ravikumar, and E.~Xing, ``Learning sparse
  nonparametric dags,'' in \emph{International Conference on Artificial
  Intelligence and Statistics}.\hskip 1em plus 0.5em minus 0.4em\relax PMLR,
  2020, pp. 3414--3425.

\bibitem{CAM}
\BIBentryALTinterwordspacing
P.~Bühlmann, J.~Peters, and J.~Ernest, ``{CAM: Causal additive models,
  high-dimensional order search and penalized regression},'' \emph{The Annals
  of Statistics}, vol.~42, no.~6, pp. 2526 -- 2556, 2014. [Online]. Available:
  \url{https://doi.org/10.1214/14-AOS1260}
\BIBentrySTDinterwordspacing

\bibitem{DataFood}
G.~Menegozzo, D.~Dall’Alba, and P.~Fiorini, ``Cipcad-bench: Continuous
  industrial process datasets for benchmarking causal discovery methods,'' in
  \emph{2022 IEEE 18th International Conference on Automation Science and
  Engineering (CASE)}.\hskip 1em plus 0.5em minus 0.4em\relax IEEE, 2022, pp.
  2124--2131.

\bibitem{Hogan2021}
A.~Hogan, E.~Blomqvist, M.~Cochez, C.~D’amato, G.~Melo, C.~Gutierrez,
  S.~Kirrane, J.~E.~L. Gayo, R.~Navigli, S.~Neumaier, A.-C.~N. Ngomo,
  A.~Polleres, S.~M. Rashid, A.~Rula, L.~Schmelzeisen, J.~Sequeda, S.~Staab,
  and A.~Zimmermann, ``Knowledge graphs,'' \emph{ACM Computing Surveys},
  vol.~54, pp. 1--37, 2021.

\bibitem{Harary1965}
F.~Harary, R.~Z. R.~Z. Norman, and D.~Cartwright, \emph{Structural models: an
  introduction to the theory of directed graphs}.\hskip 1em plus 0.5em minus
  0.4em\relax Wiley, 1965.

\bibitem{Gallian2001}
J.~Gallian, ``A dynamic survey of graph labeling,'' \emph{Electron. J. Combin.,
  Dynamic Surveys}, vol.~19, 11 2000.

\bibitem{wille2004sparse}
A.~Wille, P.~Zimmermann, E.~Vranov{\'a}, A.~F{\"u}rholz, O.~Laule, S.~Bleuler,
  L.~Hennig, A.~Preli{\'c}, P.~von Rohr, L.~Thiele \emph{et~al.}, ``Sparse
  graphical gaussian modeling of the isoprenoid gene network in arabidopsis
  thaliana,'' \emph{Genome biology}, vol.~5, no.~11, pp. 1--13, 2004.

\bibitem{saxe2020computational}
G.~N. Saxe, S.~Ma, L.~J. Morales, I.~R. Galatzer-Levy, C.~Aliferis, and C.~R.
  Marmar, ``Computational causal discovery for post-traumatic stress in police
  officers,'' \emph{Translational psychiatry}, vol.~10, no.~1, pp. 1--12, 2020.

\bibitem{PetersBook}
J.~Peters, D.~Janzing, and B.~Sch\"olkopf, \emph{Elements of Causal Inference:
  Foundations and Learning Algorithms}.\hskip 1em plus 0.5em minus 0.4em\relax
  Cambridge, MA, USA: MIT Press, 2017.

\bibitem{DirectLiNGAM}
S.~Shimizu, T.~Inazumi, Y.~Sogawa, A.~Hyv\"{a}rinen, Y.~Kawahara, T.~Washio,
  P.~O. Hoyer, and K.~Bollen, ``Directlingam: A direct method for learning a
  linear non-gaussian structural equation model,'' \emph{J. Mach. Learn. Res.},
  vol.~12, no.~0, p. 1225–1248, jul 2011.

\bibitem{DAGGNN}
\BIBentryALTinterwordspacing
Y.~Yu, J.~Chen, T.~Gao, and M.~Yu, ``{DAG}-{GNN}: {DAG} structure learning with
  graph neural networks,'' in \emph{Proceedings of the 36th International
  Conference on Machine Learning}, ser. Proceedings of Machine Learning
  Research, K.~Chaudhuri and R.~Salakhutdinov, Eds., vol.~97.\hskip 1em plus
  0.5em minus 0.4em\relax PMLR, 09--15 Jun 2019, pp. 7154--7163. [Online].
  Available: \url{https://proceedings.mlr.press/v97/yu19a.html}
\BIBentrySTDinterwordspacing

\bibitem{ShojaieKnownOrdering}
\BIBentryALTinterwordspacing
A.~Shojaie and G.~Michailidis, ``{Penalized likelihood methods for estimation
  of sparse high-dimensional directed acyclic graphs},'' \emph{Biometrika},
  vol.~97, no.~3, pp. 519--538, 07 2010. [Online]. Available:
  \url{https://doi.org/10.1093/biomet/asq038}
\BIBentrySTDinterwordspacing

\bibitem{lal2015}
M.~Lal, \emph{Neo4j Graph Data Modeling}.\hskip 1em plus 0.5em minus
  0.4em\relax Packt Publishing, 2015.

\end{thebibliography}

\end{document}